# Périphériques haptiques et simulation d'objets, de robots et de mannequins dans un environnement de CAO-Robotique : eM-Virtual Desktop

D. Chablat[1], F. Bennis[1], B. Hoessler[2], M. Guibert[2]

(1) : Institut de Recherche en Communications et
Cybernétique de Nantes
1, rue de la Noë
44321 Nantes France
Téléphone 02 40 37 69 54 /Fax 02 40 37 69 54
E-mail : Damien.Chablat@irccyn.ec-nantes.fr

(2) : Tecnomatix France
Espace Jouy Technologie
21, rue Albert Calmette
78353 Jouy-en-Josas France

E-mail : bhoessler@tecnomatix.fr

**Résumé :** Le travail présenté dans ce papier est relatif à l'intégration d'une interface à retour d'efforts dans un environnement de simulation. Après une analyse de l'existant dans le domaine, des conditions d'intégration et des applications potentielles, nous présentons la solution retenue et les étapes de développement. Les éléments de modélisations nécessaires à la manipulation d'objets, de mannequins et de robots à l'aide du Phantom desktop sont particulièrement développés dans ce papier. Nous montrerons également comment le retour d'effort est géré par l'application.
Cette étude a été concrétisée par le développement du module eM-Virtual Desktop associé au produit eM-Workplace de la société Tecnomatix. Le module développé est testé actuellement par un constructeur automobile européen.

**Mots clefs :** Retour d'efforts, Robot, Mannequin, Simulation.

**Abstract:** This paper presents the development of a new software in order to manage objects, robots and mannequins in using the possibilities given by the haptic feedback of the Phantom desktop devices. The haptic device provides 6 positional degree of freedom sensing but three degrees force feedback. This software called eM-Virtual Desktop is integrated in the Tecnomatix's solution called eM-Workplace. The eM-Workplace provides powerful solutions for planning and designing of complex assembly facilities, lines and workplaces. In the digital mockup context, the haptic interfaces can be used to reduce the development cycle of products. Three different loops are used to manage the graphic, the collision detection and the haptic feedback according to theirs own frequencies. The developed software is currently tested in industrial context by a European automotive constructor.

## 1- Introduction

La phase de développement est une période cruciale dans le cycle de création des produits et des services industriels. Les objectifs visés sont antagonistes : optimisation des coûts, amélioration de la qualité et des performances et réduction des délais de développement. L'ingénierie intégrée permet de répondre à ces objectifs en parallélisant les tâches et en intégrant, à tous les niveaux de développement, toutes les considérations relatives au cycle de vie du produit. Les logiciels de simulation contribuent à répondre, en partie, à ces objectifs. Ils permettent d'explorer un espace de configurations suffisamment large en un temps minimum et hors ligne de production. La réalisation de prototype de produit est alors réduite au strict nécessaire. La définition de la tâche et du poste de travail est optimisée en intégrant tous ces aspects le plus tôt possible dans le cycle de création du produit.

La définition de la maquette numérique des éléments nécessaires à la simulation est réalisée à l'aide de logiciels de CAO. Une grande partie des aspects géométriques centrés produits est déjà gérée par les applications de simulation de mécanismes, robots, machines, mannequin, de poste de travail,... Par contre, l'interface utilisateur est limitée, la plupart du temps, aux aspects graphiques des icônes et des boîtes de dialogues. La maturité et la grande diffusion de ces logiciels sont certainement à l'origine de l'émergence d'un nouveau besoin. Il s'agit des fonctions centrées sur le point de vue de l'ingénieur ou de l'opérateur qui participe à la définition et à la création de produit ou du service. Parmi ces fonctions, l'immersion, la présence, la navigation, l'interaction sont identifiées comme les caractéristiques clefs des outils de la réalité Virtuelle. Celle-ci est actuellement en plein développement et les résultats de recherche commencent progressivement à quitter les centres de recherche et de développement pour être déployés dans le cadre de projets industriels [1, 2].

La littérature générale sur le thème de la réalité virtuelle est relativement abondante [3, 4]. Nous nous limitons dans ce papier au développement de notre étude spécifique en insistant sur les points particuliers de notre approche par rapport à une approche générale.

La stratégie générale utilisée dans les logiciels de la réalité virtuelle existants peut être résumée par les étapes suivantes :

- <u>Récupération</u> de la maquette numérique des éléments de la scène qui peuvent provenir de plusieurs applications CAO différentes. Les éléments de la scène sont transformés en un format unique.
- <u>Facétisation</u> et tesséllation des éléments. Cette opération a pour but de réduire la taille occupée par des données géométriques manipulées afin de respecter la contrainte principale relative au temps réel.
- <u>Simulation</u> et recherche de solution pour la tâche désirée.
- <u>Modification et réactualisation des modèles</u>. Après la simulation, l'utilisateur doit prendre en compte les modifications éventuelles suggérées par la simulation et réactualiser le modèle CAO en conséquence.

Chaque étape, de cette boucle simplifiée, présente des difficultés et des contraintes supplémentaires pour le développement du produit. La phase de simulation est assez critique pour notre application car le logiciel de simulation de robots dispose de fonctions dédiées indispensables à la simulation de robots. Il n'est pas possible de les reproduire dans le système de réalité virtuelle. De plus, la recherche des collisions entre les objets de l'environnement virtuel, est rendue plus difficile à cause de la dégradation des volumes en facettes. Notons que la réalisation des autres étapes peut être très longue et difficile à réaliser lorsque les partenaires du projet sont situés sur des sites distants et utilisent des maquettes numériques d'origines différentes [1, 5].

## 2- Description du contexte et des spécifications de l'étude

Notre étude porte sur le développement d'une application informatique permettant l'intégration d'un périphérique haptique dans un environnement de simulation. On s'intéresse, plus particulièrement, aux fonctions relatives à la manipulation d'objets, robots ou de mannequins dans l'environnement encombré d'un poste de travail. Les applications potentielles du produit sont : la simulation pour l'optimisation du poste de travail, la génération de trajectoire, la formation des opérateurs, l'entraînement à des tâches difficiles, la maintenance, et l'analyse ergonomique du poste de travail.

Pour l'entreprise Tecnomatix, le choix du périphérique devait être guidé par les contraintes de coût, de facilité d'utilisation et de mise en œuvre. L'outil ne doit pas être encombrant, ce qui élimine les solutions de type exo-squelette, casque,…. Il doit être capable de s'adapter à une large palette de robots, et d'objets manipulés. Il doit être intégrable directement dans l'application de simulation de robots tout en conservant, pour l'utilisateur, une fluidité des mouvements. Les contraintes du temps réel doivent être contrôlées.

Toutes ces contraintes sont antagonistes et nécessitent une stratégie globale d'harmonisation des fréquences pour le traitement des données relatives à la visualisation, à l'interface haptique, au calcul, etc...

## 3- Analyse fonctionnelle

Il existe plusieurs méthodes, automatiques ou semi-automatiques, pour la détection et l'évitement de collision en génération de trajectoire. Ces méthodes ne sont pas encore suffisamment matures pour une intégration complète et un respect de la contrainte du temps réel.

Nous nous intéressons principalement à l'approche de génération de trajectoire effectuée par un opérateur expert en s'appuyant sur des fonctionnalités interactives. Dans les logiciels existants, l'opérateur dispose de plusieurs fonctions élémentaires basées sur l'utilisation de la souris 2D ou 3D de type "*space mousse*". Il est souvent guidé par des boîtes de dialogue et des menus déroulants. Compte tenu de la complexité géométrique sous-jacente à ces problèmes de simulation, les interfaces graphiques ne sont pas suffisamment naturelles pour traiter rapidement et efficacement les situations difficiles. On retrouve également des fonctions plus évoluées qui affichent les entités et les zones en collision, ou encore les fonctions de type "*stop sur collision*". Ces dernières fonctions arrêtent le mouvement de l'objet manipulé lorsque la collision est détectée, sans contrôler le mouvement de la souris. L'opérateur est averti par un simple signal sonore et un changement de couleurs.

Pour fixer les idées, prenons l'exemple d'un robot à structure parallèle qui évolue dans un environnement encombré en manipulant un objet de forme complexe. Lorsqu'une collision est détectée, la solution pour se dégager n'est pas toujours simple. Les déplacements de la souris ne sont pas en adéquation avec les mouvements des objets manipulés dans l'espace à six dimensions en position et orientation. Dans certaines situations, l'opérateur doit utiliser plusieurs étapes pour réaliser un mouvement élémentaire du robot. Une méthode, qui aide l'opérateur à faire glisser des objets sur les frontières de l'environnement, est proposée dans [6]. Dans ce cas, la direction de glissement n'est pas perçue par l'utilisateur.

D'une manière générale, les contraintes imposées par la souris (2D ou 3D) et par l'interface graphique ne sont pas très naturelles pour l'expert. Les périphériques haptiques doivent apporter des éléments de réponse à ce problème.

## 4- Périphériques haptiques

Les périphériques de l'ingénierie virtuelle se développent et s'adaptent de plus en plus aux besoins spécifiques [4]. La possibilité d'intégrer le retour d'effort par l'intermédiaire du périphérique haptique offre une souplesse supplémentaire par rapport aux périphériques classiques.

Par exemple, pour ouvrir la porte d'une voiture, l'opérateur doit être guidé par le périphérique et contraint à effectuer un mouvement circulaire dans un plan le plus naturellement possible. Il devra également ressentir la butée de la porte voire sa résistance. Le retour d'effort reproduit, assez fidèlement par rapport à la réalité, la notion du mouvement temporel et spatial (position et orientation). Grâce à ce type d'interactions, l'expert a le temps et la possibilité de se concentrer mieux sur les fonctions de la tâche à définir. Il s'encombre moins des contraintes imposées par les menus et les boîtes de dialogues.

### 4.1- Retour d'effort

Il est possible de classer l'application du retour d'effort en trois classes. Dans la première classe, l'effort retourné est constant quelle que soit la situation de collision rencontrée. Il informe uniquement sur la présence ou non de celle-ci en "bloquant" les mouvements du périphérique ; cela donne la sensation d'une frontière. Le choix de la valeur de l'effort n'est pas important, car l'opérateur a seulement besoin de distinguer entre les mouvements libres (sans retour d'effort virtuel) et les mouvements qui rencontrent des obstacles. Dans la deuxième classe, l'effort est variable en fonction de l'importance et du type de collision. Il est également possible de distinguer la masse des objets manipulés et d'autres types de critères de différentiation. La palette des valeurs des efforts virtuels retournés doit être suffisamment large pour pouvoir distinguer les objets manipulés. La troisième prend en compte la dynamique du mouvement des objets manipulés [7].

Il existe également des périphériques qui retournent un couple mais qui nécessitent une fraction supplémentaire du temps de traitement. Leur coût est également un frein qui ne favorise pas leur diffusion.

### *4.2- Espace de la scène et espace de déplacement*

En général, l'espace accessible du périphérique haptique est relativement faible par rapport à la taille de l'espace de la scène simulé. Il existe plusieurs problèmes relatifs à la correspondance entre les déplacements dans ces deux espaces.

- Facteur d'échelle : L'application d'un facteur d'échelle doit s'adapter à la précision souhaitée du mouvement. Par exemple, pour des opérations d'assemblage de précision, une bonne dextérité de l'opérateur est requise. Le mouvement des entités dans l'espace virtuel doit donc être aussi précis. Inversement, lors de la réalisation de mouvement de transfert d'objet, l'amplitude de mouvement dans l'espace virtuel peut être plus grande.
- Espace accessible : Lorsque les mouvements générés dans l'espace virtuel sont de faible amplitude, il n'est pas possible de générer des mouvements dans toute la fenêtre graphique. Une stratégie doit donc être mise en place pour changer l'origine des déplacements.
- Direction de déplacement : Le repère du périphérique peut être associé à celui de la vision (point de vue de la scène) ou celui du repère de la tâche. Ce problème est lié à la difficulté de génération de mouvements, par l'utilisateur, dans l'espace virtuel à 6 degrés de liberté (translation et rotation).

### *4.3- Gestion du contact / collision*

La génération de retour haptique nécessite de nombreux tests de collision entre l'entité manipulée et son environnement. Le calcul doit être rapide et doit retourner les informations nécessaires à la génération du retour haptique (effort et couple).

Pour réaliser la détection de collision, il existe plusieurs outils, comme la librairie « contact », développé par l'INRIA [8] ou l'application Voxel PointShell (VPS) [9]. Cette dernière est actuellement intégrée par SensAble dans les dernières versions de GHOST sous Windows [10].

Le calcul théorique permet la génération d'une force et d'un couple, en utilisant les coordonnées du point de contact et la normale à la surface rencontrée en ce point ainsi que le centre virtuel de rotation, par exemple, le centre de la main de l'utilisateur.

## 5- Structure et fonctionnalités de l'application eM-Virtual Desktop

### *5.1- Présentation générale*

Deux périphériques haptiques sont intégrés dans l'application eM-Virtual Desktop. Le premier est un Phantom Desktop (*Fig. 1*), périphérique haptique de la société SensAble Technologies [11] et le second, un gant de données CyberGlove (ou gant numérique) (*Fig. 2*), de la société Virtual Technologies [12]. Dans cet article, seulement l'intégration du Phatom Desktop est présentée.

Nous avons choisi le Phantom en raison de la similitude qu'il présente avec les éléments que nous traitons dans le cadre de ce projet. Sa structure articulaire, équivalente à un robot, permet à l'expert de manipuler les éléments du poste de travail avec des mouvements assez naturels.

Tous les produits Phantom de la société SenAble Technologies possèdent la même architecture de robot. Le déplacement du Phantom est réalisé grâce à un stylet. Sur son extrémité est placé un petit bouton. Le déplacement du stylet peut s'apparenter au déplacement d'un crayon dans l'espace (même prise en main). Le périphérique utilisé dans notre application est le Phantom Desktop. Son espace de travail plus grand couvre une zone de 16 cm x 13 cm x 13 cm. Les six degrés de liberté de mouvement sont mesurés et renseignent sur la position des entités manipulées. La résolution de la mesure de la position est 0.02 mm. Par ailleurs, les forces engendrées par ce périphérique s'appliquent sur un point. Il n'est pas possible mécaniquement de générer un couple sur la main de l'utilisateur. Le retour d'effort est renseigné à l'aide de trois degrés de liberté : 6.4 N maxi et 1.4 N en continu (pas de retour en couple).

Pour les domaines d'application qui nécessitent un retour haptique sur les six degrés de liberté, il existe le Phantom 1.5/6DOF. Cependant, il n'a pas été retenu par la société Tecnomatix pour être intégré dans cette application.

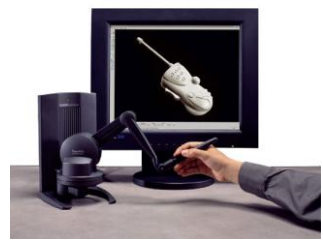 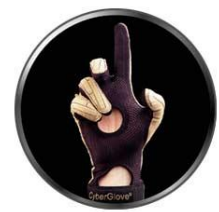

*Fig. 1* : **Phantom desktop**     *Fig. 2* : **Cyberglove**

L'application eM-Virtual Desktop est complètement intégrée dans l'application eM-Workplace. Pour en simplifier l'apprentissage, son démarrage s'assimile au démarrage de n'importe quel module et conserve la même charte graphique (*Fig. 3*). L'application regroupe les deux interfaces haptiques. L'interface est identique et l'utilisateur choisit en fonction de son besoin (*Fig. 4*).

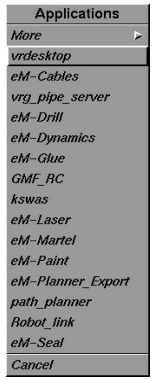
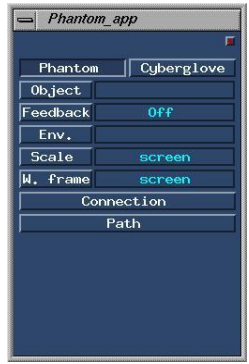
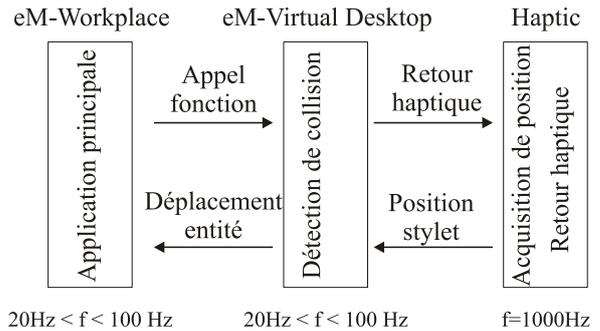

***Fig. 3* : Liste des applications disponibles dans eM-Worksplace**

***Fig. 4* : Fenêtre principale de l'application eM-Virtual desktop**

***Fig. 5* : Schéma de communication entre les processus de l'application**

### 5.2- Intégration dans eM-Worksplace

Le principal problème rencontré lors de l'intégration du périphérique haptique dans un environnement de CAO-Robotique est lié à la communication entre plusieurs programmes possédant des fréquences différentes. Le contrôle d'un périphérique haptique se ramène à la gestion de processus avec des fréquences proches du temps réel. En effet, l'affichage des objets en mouvement peut être échantillonné de manière convenable autour de 10 Hz. Par contre, les retours d'effort doivent être réalisés à des fréquences beaucoup plus importantes. Pour le contrôle du Phantom Desktop, la fréquence d'échantillonnage de la boucle de contrôle est de 1000Hz.

La communication entre la boucle haptique et notre application est réalisée grâce à un "socket" qui réalise une liaison de type client / serveur. Nous avons un serveur associé à la boucle haptique, et un client associé à la boucle d'événement du processus eM-Virtual Desktop. Pour que la communication n'interrompe pas la boucle haptique, le socket doit être *non-bloquant*. La communication entre les applications est la suivante (***Fig. 5***) :

- eM-Worksplace est l'application maître. L'application eM-Virtual Desktop est alternativement activée.
- Si le retour haptique est activé, une requête est réalisée sur le serveur pour connaître la position du stylet.
- Pour chaque nouvelle situation d'un objet déplacé, on teste l'éventualité d'une collision avec l'environnement. Le cas échéant, une force est renvoyée sur le stylet, sinon la nouvelle situation est actualisée dans eM-Workplace. Entre deux requêtes du client vers le serveur, la boucle haptique contrôle le retour d'effort en assurant la cohérence avec la position courante du stylet.

Précisons, par ailleurs, que la communication à l'aide du socket permet de connecter deux processus présents sur la même station de travail ou sur deux stations séparées grâce à l'utilisation du réseau TCP-IP. Ainsi, il est possible de relier deux systèmes d'exploitation différents ou de distribuer les calculs sur plusieurs processeurs. Les tests ont été réalisés à l'IRCCyN sur une station SGI OCTANE, de type bi-processeur Mips R12000 à 275MHz.

### 5.3- Détection de collision et génération du retour haptique

Pour les contraintes de simplifications et de robustesses données précédemment, nous limitons notre recherche à la détermination de l'effort. Nous laissons également la possibilité, à l'utilisateur, de désactiver la génération de retour haptique pour permettre une utilisation du stylet comme une souris 3D.

Pour le calcul d'interférence dans eM-Virtual Desktop, nous avons exploité directement les fonctions disponibles dans l'application eM-Workplace. Ainsi, il n'existe aucune perte d'information et la mise à jour de la base de données est complètement intégrée.

Pour chaque situation de l'entité manipulée dans son environnement, la fonction d'interférence retourne les coordonnées des deux points les plus proches entre les deux entités. Elle retourne également la distance entre ces entités lorsqu'il n'y a pas de collision, et zéro sinon.

Toujours pour des raisons de réduction du temps de calcul, nous évitons le calcul des coordonnées du point de contact lors de la détection de collision. Des stratégies jugées coûteuses en temps de calcul pour notre application sont disponibles dans la librairie Contact. Notre approche de détection de collision est basée sur l'exploitation d'une zone de sécurité autour de l'environnement virtuel. Ainsi, nous pouvons générer une sensation de contact pour l'utilisateur en définissant un point de contact et une normale à la surface de contact grâce aux positions de ces deux points.

### 5.4- Déplacement des solides

Les axes de déplacement des solides, via le déplacement du stylet de l'utilisateur, sont définis de trois manières différentes. La première semble être la plus naturelle. Elle correspond aux axes Y et Z du plan de l'écran et l'axe X normal à ce plan. Pour ce faire, le Phantom Desktop doit être placé juste à côté de l'écran (***Fig. 6***).

La seconde suit les axes du repère principal de la cellule. Son utilisation est limitée car lorsque l'on déplace le point d'observation de l'utilisateur sur la cellule, les déplacements ne sont pas intuitifs. La dernière suit des axes de déplacement, définis par l'utilisateur.

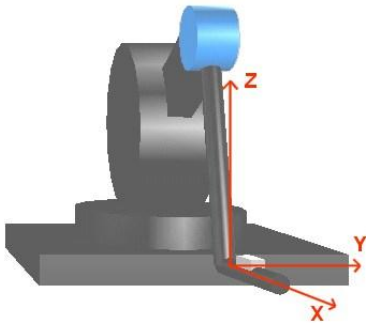

**Fig. 6 : Les axes du Phantom**

*5.5- Facteur d'échelle*

Pour permettre de résoudre les problèmes de changement d'échelle entre la dimension de la cellule de travail et l'espace accessible du Phantom Desktop, quatre niveaux de sensibilité sont définis pour les mouvements de translation (Rough, medium, fine et screen ). Le niveau screen définit un facteur d'échelle adaptatif, entre la taille de l'environnement affiché à l'écran et l'espace accessible par le Phantom Desktop. La valeur de l'échelle est définie par l'utilisateur en fonction du grossissement (« zoom+/zoom- »).

Les valeurs numériques des seuils peuvent être modifiées par l'utilisateur. Ces valeurs se trouvent dans les fichiers de configuration du logiciel.

*5.6- Environnement associé à la détection de collision*

Pour simplifier les calculs, il n'est pas nécessaire de tester toutes les entités de l'environnement ainsi que tous ses détails. En effet, le temps de calcul des collisions dépend fortement du nombre d'entités à tester. L'utilisateur doit utiliser une stratégie de découpage de l'environnement en groupe de test.

Une fonctionnalité d'eM-Worksplace permet de définir un groupe d'entités qui doit être testé avec un autre groupe. Les groupes sélectionnés peuvent être modifiés lors du déplacement de l'objet. La fonction retourne la distance minimale ou la présence éventuelle d'une collision.

## 6- Entités manipulées

Trois types d'entités peuvent être manipulés via le stylet du Phantom Desktop dans une cellule eM-Worksplace. Pour chaque type (solides, robots et mannequins), un comportement mécanique est associé. En sélectionnant une entité dans la cellule, le système associe automatiquement son modèle cinématique et ses propriétés de déplacement. La programmation en C++ de l'application autorise facilement l'ajout d'entités supplémentaires. Le bouton du stylet permet d'activer ou désactiver la sélection et le déplacement des objets.

*6.1- Manipulation d'un solide*

La manipulation d'un solide via un périphérique de réalité virtuelle est une tâche complexe car la manipulabilité de la main sur le stylet du Phantom Desktop est différente de celle du contact direct avec la main. En particulier, nous devons définir le centre de rotation de la pièce en faisant correspondre le repère en ce point à celui de notre main.

Il est possible d'utiliser trois repères dans notre application. Le premier est défini par construction, c'est le « self origin ». Pour un cube, par exemple, il est placé au centre de sa face inférieure. Le second est le « geometric center » qui permet des manipulations plus intuitives. Pour finir, si ces deux repères sont insuffisants, l'utilisateur peut utiliser sa propre définition (*Fig. 7*).

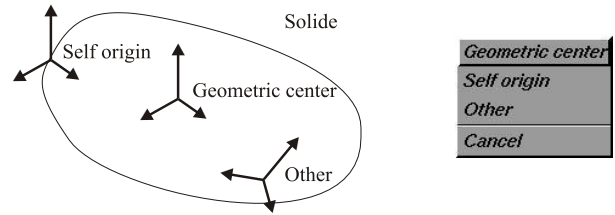

**Fig. 7 : Repères associés et fenêtre de dialogue pour le déplacement de solides**

*6.2- Robot*

Un robot est composé de plusieurs corps solides liés les uns aux autres par des articulations. La situation relative des corps est donc variable pendant la simulation.

En principe, le mouvement du stylet peut être associé à n'importe quel solide du robot. Pour simplifier, nous avons conservé uniquement la possibilité de manipuler la base ou l'effecteur. Si l'on sélectionne l'option « base », alors le déplacement est équivalent au déplacement d'un corps rigide (*Fig. 8*). Par contre, si l'on choisit l'option effecteur « Tcpf », alors le modèle géométrique inverse du robot est utilisé. Ce modèle calcule les coordonnées articulaires à partir de la position et de l'orientation du robot. Il retrouve donc les situations relatives des corps du robot. Les limites de l'espace de travail peuvent être considérées comme un obstacle.

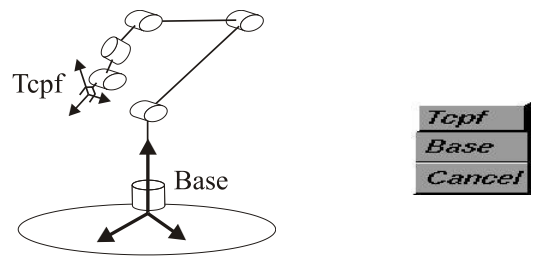

**Fig. 8 : Repères de déplacement et fenêtre de dialogue d'un robot**

Grâce à l'intégration totale d'eM-Virtual Desktop dans eM-Workplace, il est possible de traiter n'importe quel robot de la bibliothèque disposant d'un modèle géométrique inverse (*Fig. 9*).

Lorsqu'il existe plusieurs solutions au modèle géométrique inverse, la solution retenue est la plus proche de la situation précédente. Cela permet d'assurer la continuité de la trajectoire dans l'espace articulaire. Cependant, il est possible de changer la solution retenue en utilisant les fonctionnalités d'eM-Workplace.

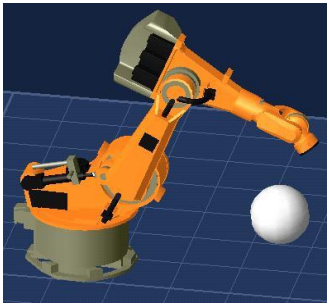

*Fig. 9* : **Robot dans eM-Virtual Desktop**

### 6.3- *Manipulation d'un mannequin*

La dernière entité manipulée est un mannequin. Avec le stylet du Phantom Desktop, il est possible de déplacer, tout le mannequin, une des deux mains ou les deux mains à la fois en utilisant les modèles géométriques du mannequin (*Fig. 10*). Le mannequin disponible possède 56 degrés de mobilité (*Fig. 11*).

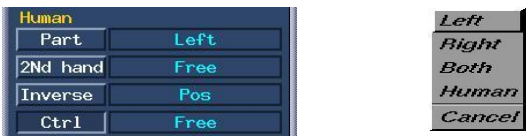

*Fig. 10* : **Fenêtre de dialogue associé au déplacement d'un mannequin**

Des contraintes peuvent être ajoutées au mannequin, pour bloquer le déplacement du tronc.

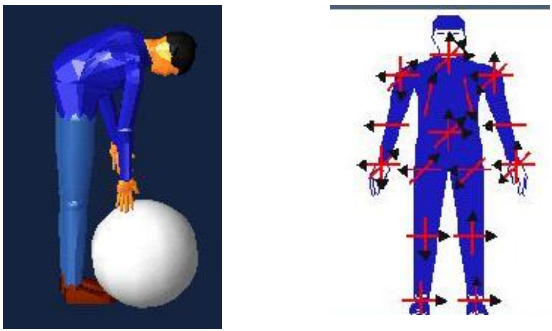

*Fig. 11* : **Un mannequin dans eM-Virtual Desktop avec ses degrés de mobilité**

### 6.4- *Génération de trajectoires*

La génération de trajectoire doit permettre de mémoriser le déplacement des entités manipulées. Sa création peut se faire soit en manuel, soit en automatique. Dans ce cas, les repères constituant la trajectoire sont créés suivant des intervalles de temps ou de distances. Le lissage de la trajectoire enregistré par l'utilisateur n'est pas généré par l'application.

## 7- Conclusion

Malgré les contraintes imposées par l'intégration du périphérique haptique dans l'environnement de simulation de CAO-Robotique, les résultats obtenus sont très satisfaisants. Des tests supplémentaires relatifs à ce développement sont actuellement réalisés par un constructeur automobile européen. Ces tests doivent évaluer le module dans le cadre d'une application industrielle et définir les futures voix de développement. Il sera possible de coupler notre solution avec les méthodes proposées dans [6, 13].

## 8- Remerciements

## 9- Bibliographie